%% file: acl_latex_preprint.tex
\documentclass[11pt]{article}

\usepackage[preprint]{acl}

\usepackage{times}
\usepackage{latexsym}

\usepackage[T1]{fontenc}

\usepackage[utf8]{inputenc}

\usepackage{microtype}

\usepackage{inconsolata}

\usepackage{graphicx}
\usepackage{amsmath}
\usepackage{booktabs}
\usepackage{float}
\usepackage{xcolor}
\usepackage{tikz}
\usetikzlibrary{positioning,arrows.meta,fit,backgrounds,calc,
                decorations.pathreplacing}
                
\usepackage{stfloats}

\title{One Policy Is Enough: Single-Agent Reinforcement Learning Outperforms
Tree Search for Chemistry Tool Learning}

\author{
  \textbf{Armin Dariani\textsuperscript{1,2}},
  \textbf{Sifan Wu\textsuperscript{1,2}},
  \textbf{Bang Liu\textsuperscript{1,2*}},
  \textbf{Entao Yang\textsuperscript{3*}}
  \\
  \textsuperscript{1}DIRO, Université de Montréal
  \\
  \textsuperscript{2}Mila - Quebec Artificial Intelligence Institute
  \\
  \textsuperscript{3}Innovation Campus Delaware, Air Liquide
  \\
  \small{
    \href{mailto:armin.zolfagharidariani@mila.quebec}
    {armin.zolfagharidariani@mila.quebec},
    \href{mailto:sifan.wu@umontreal.ca}
    {sifan.wu@umontreal.ca}
  }
  \\
  \small{
    \href{mailto:bang.liu@umontreal.ca}
    {bang.liu@umontreal.ca},
    \href{mailto:entao.yang@airliquide.com}
    {entao.yang@airliquide.com}
  }
  \\
  \small{\textsuperscript{*}Corresponding authors}
}

\begin{document}
\maketitle
\begin{abstract}
Chemistry questions often demand exact computation and database lookups that a language model cannot supply from its parameters, so it must reach for external tools. Tool use here is a three-part problem: select the right tool from a large pool, fill it with correctly typed arguments, and chain calls so that each consumes the outputs of the last. CheMatAgent, a previously published system, addresses this with hierarchical evolutionary MCTS: separate policy and execution models searching tool-call trees under two learned critics, one regressed partly onto GPT-assigned scores. \textbf{We show that a single policy suffices.} Our model interleaves reasoning, tool calls, and returns in one left-to-right generation, trained by a supervised warm-up and then outcome-level reinforcement learning against a programmatic reward read directly off the gold call chain, which leaves no learned critic and no judge in the training loop. On ChemToolBench multiple-tool comprehensive chemistry, on both backbones CheMatAgent use, we improve Tool F1 by $5.5\%$ and Return F1 by $9.6\%$ on Qwen-2.5-7B, and by $3.7\%$ and $3.9\%$ on Llama-3.1-8B, compared with their strongest search configuration, at one model invocation per question, against a search whose cost grows with the tree; we also lead answer Pass Rate on Qwen-2.5-7B.
\end{abstract}

\section{Introduction}
\label{sec:intro}
\input{fig_protocol}

Large language models are increasingly capable on chemistry problems such as molecule generation and reaction prediction \citep{zhang2024chemllm}, but their parameters cannot store every measured property, and they cannot carry out the exact calculations many questions require. The practical remedy is to let the model call external tools as it reasons: cheminformatics libraries, property predictors, database lookups, as in ChemCrow \citep{bran2023chemcrow} and CACTUS \citep{mcnaughton2024cactus}.

Building such an agent reliably remains challenging for several reasons. \textbf{First, chemistry tools are highly specialized and challenging for general-purpose LLMs to reason.} Modern chemistry agents interact with a large collection of tools covering molecular property prediction, reaction analysis, quantum chemistry, database retrieval, and materials simulation. Many tools expose similar interfaces while differing subtly in their underlying assumptions, input requirements, or prediction targets, making accurate tool selection a non-trivial reasoning problem rather than simple retrieval. \textbf{Second, chemistry tools require structured inputs with strict scientific validity and computational precision.} Their arguments often consist of formal chemical representations, such as SMILES strings, molecular identifiers, chemical formulas, or reaction specifications, where syntactic or semantic errors can invalidate execution or lead to scientifically incorrect results. Generating such arguments therefore requires precise structured prediction grounded in chemical knowledge.
\textbf{Third, chemistry problems often require long-horizon tool interaction instead of isolated function calls.} Solving a task may involve retrieving molecular information, predicting intermediate properties, and using those outputs to parameterize subsequent computations. Because each tool call depends on the correctness of previous ones, early mistakes propagate through the execution trajectory and compound into downstream failures.


The current state-of-the-art system on ChemToolBench, CheMatAgent's Hierarchical Evolutionary Monte-Carlo Tree Search (HE-MCTS), meets this with machinery. One model plans tool selection while a second fills parameters; a Process Reward Model and an Outcome Reward Model are trained as search critics, the latter regressed onto a mixture of rule-based and GPT-assigned scores; the policy is then fine-tuned on trajectories the search collected. The results are strong and the pipeline is expensive: four models to train and coordinate, plus a per-query search cost the authors themselves note as a drawback. That leaves open whether any of it is necessary.

Recent advances in reinforcement learning suggest the latter may be possible. Reinforcement learning with verifiable rewards can elicit strong reasoning in a single model \citep{deepseekai2025r1}, and the idea extends to tool use \citep{feng2025retool,jin2025searchr1}. Following that line, we train one policy that alternates thinking and tool calls inside a single generation and optimize it end to end with Group Relative Policy Optimization \citep{shao2024deepseekmath} against a reward computed directly from the gold call chain. The model reasons, calls a tool, reads the real execution result, and continues until it answers (Figure~\ref{fig:protocol}); there is no execution model, no learned critic, and no search. To keep the comparison controlled we reuse CheMatAgent's benchmark, toolpool, and candidate-tool retrieval, and vary only how the model is trained.

\paragraph{Contributions.}
\textbf{(i)} We recast chemistry tool learning as single-agent reinforcement learning, in which one policy interleaves reasoning with live tool execution in a single generation, a procedure we call our \emph{rollout protocol} (\S\ref{sec:rollout}). It is trained with GRPO against a verifiable reward, dispensing with the execution model, the PRM and ORM critics, and the tree search of the previous state of the art \citep{wu2025CheMatAgent}. \textbf{(ii)} We develop a systematic reward framework for multi-tool calling, introducing programmatic rewards that capture answer correctness, tool selection, tool-argument accuracy, and their combinations. We further analyze the trade-offs and failure modes induced by different reward designs, including tool spamming and non-termination. \textbf{(iii)} 
Our single-policy model consistently outperforms existing SOTA approaches across multiple metrics. It improves Tool F1 by $5.5\%$ and Return F1 by $9.6\%$ on Qwen-2.5-7B, and by $3.7\%$ and $3.9\%$ on Llama-3.1-8B, respectively. To disentangle model improvements from evaluation effects, we further evaluate each untrained backbone under the same rollout protocol, as untrained performance reflects both the evaluation framework and the model capability. Compared with this baseline, supervised warm-up provides a larger contribution than reinforcement learning alone, while reinforcement learning primarily improves tool precision at the expense of recall.

\section{Related Work}
\label{sec:related}

\paragraph{Tool learning: data and methods.}
Toolformer \citep{schick2023toolformer} showed that a model can learn on its own when to call an API, and ToolLLM \citep{qin2023toolllm} scaled this to over $16{,}000$ real APIs with the ToolBench benchmark. Later resources target the two hard parts, choosing a tool and filling its arguments: API-Bank \citep{li2023apibank}, Seal-Tools \citep{wu2024sealtools}, and ToolACE \citep{liu2024toolace}, which synthesises function-calling data with a multi-agent pipeline. The Berkeley Function Calling Leaderboard \citep{patil2025bfcl} is the standard evaluation, now extended to multi-turn, stateful settings. Most of this line teaches tool use by supervised fine-tuning alone. We use SFT only as a warm-up before online RL in which the model calls the tools during training.

\paragraph{LLM agents for chemistry.}
Science suits tool-augmented agents because answers must be precise and checkable. ChemCrow \citep{bran2023chemcrow} combines eighteen expert tools to plan organic syntheses, CACTUS \citep{mcnaughton2024cactus} pairs open LLMs with cheminformatics packages such as RDKit, SciAgent \citep{ma2024sciagent} extends tool-augmented reasoning across scientific benchmarks, and ChemLLM \citep{zhang2024chemllm} instead builds chemical knowledge in through pretraining. Closest to us is CheMatAgent \citep{wu2025CheMatAgent}, the strongest of these, contributing a $137$-tool chemistry pool, the ChemToolBench dataset, and HE-MCTS: a policy model and an execution model are kept separate, a trained Process Reward Model and Outcome Reward Model guide tree search, and the policy is fine-tuned step by step on search trajectories. Accuracy is strong, but the pipeline is large and costly at inference, a point the authors themselves raise. We keep their benchmark and toolpool unchanged and ask whether much simpler training reaches the same level.

\paragraph{Reinforcement learning for reasoning and tool use.}
DeepSeek-R1 \citep{deepseekai2025r1} drew strong reasoning out of a single model almost entirely by reinforcement learning with verifiable rewards, using GRPO \citep{shao2024deepseekmath}, which drops PPO's learned critic and scores each sample against its group mean. The recipe then moved to tool use: ReTool \citep{feng2025retool} interleaves code execution with reasoning, and Search-R1 \citep{jin2025searchr1} does the same for search, masking retrieved tokens to keep training stable, a device we reuse. Closest to our reward design, ToolRL \citep{qian2025toolrl} decomposes the reward over tool names, parameter names and parameter values rather than matching the answer, Nemotron-Research-Tool-N1 \citep{zhang2025nemotron} uses a binary format-and-correctness reward on largely single-turn benchmarks, and MatchTIR \citep{qu2026matchtir} adds turn-level rewards from matched tool traces. We share this recipe and differ in setting and in signal. A chemistry question needs several dependent calls from a large specialized pool, each argument a SMILES string or formula, rather than one code block or one query, and our reward is read off the gold call chain, giving partial credit over tool names and their arguments rather than scoring the answer.

\section{Task and Benchmark}
\label{sec:task}

\subsection{Problem Formulation}
\label{sec:problem} 
Given a chemistry question $q$ in natural language, an agent must produce a final answer by calling tools from a fixed pool $\mathcal{T}=\{t_1,\dots,t_m\}$, each a Python function exposed through an OpenAI-style JSON signature. Most questions cannot be answered with a single call: they require a chain in which a later call, and the arguments it takes, depend on the values earlier calls returned. The agent must therefore select the right tools \emph{and} order dependent calls correctly, and it is judged both on how well its calls match a reference and on whether its final answer is correct. Every question is annotated with a validated ground-truth calling chain $C=[(t_i,\text{args}_i,r_i)]_{i=1}^{L}$ recording the tools, their arguments, and their returns; we use $C$ both to build supervised targets and to compute rewards.

\subsection{ChemToolBench}
\label{sec:chemtoolbench}
We use ChemToolBench, the tool-learning benchmark released with CheMatAgent \citep{wu2025CheMatAgent}. It has a chemistry and a materials split, each with single-tool and multi-tool questions partitioned into train, development, and test, and every question is paired with a validated gold calling chain. The chemistry split has $10{,}441$ single-tool and $2{,}023$ multi-tool questions, divided train/development/test as $8353/1044/1044$ and $1623/200/200$. The materials split is comparable in size ($15{,}742$ and $1{,}623$).

Our experiments target the harder \emph{multiple-tool} chemistry split, whose questions chain between two and six dependent calls, $3.15$ on average. We evaluate on all $200$ test questions and train on the $1566$ of $1623$ training questions whose gold chains still reproduce against the current tools (Appendix~\ref{app:refresh}). We keep the benchmark and its tools otherwise unchanged and vary only how the model is trained, so any difference in performance is attributable to the training method rather than to the data or the tools.

\subsection{Chemistry Tool Pool}
\label{sec:toolpool}
The chemistry split is served by tools drawn from four libraries (Table~\ref{tab:toolpool}) and rewritten by \citet{wu2025CheMatAgent} into a uniform, self-describing format. They span cheminformatics utilities, property and structure predictors, reaction and stoichiometry calculators, and drug-likeness scorers, and each is invoked by its library-qualified name (\texttt{<lib>/<func>}) with typed JSON arguments. Some run offline; others call live web services, which is what makes a small number of released gold return values go stale over time (Appendix~\ref{app:refresh}).

\begin{table}[t]
\centering
\small
\begin{tabular}{lr}
\toprule
\bf Source & \bf Amount \\
\midrule
ChemCrow\footnotemark[1]        & $8$ \\
CACTUS\footnotemark[2]          & $10$ \\
chemlib\footnotemark[3]         & $24$ \\
pymatgen\footnotemark[4]        & $82$ \\
Chemistry Tools\footnotemark[5] & $13$ \\
\bf In total: & $137$ \\
\bottomrule
\end{tabular}
\caption{The tool pool by source, counted after organization and rewriting, reproduced from \citet{wu2025CheMatAgent}.}
\label{tab:toolpool}
\end{table}
\footnotetext[1]{\url{https://github.com/ur-whitelab/chemcrow-public}}
\footnotetext[2]{\url{https://github.com/pnnl/cactus}}
\footnotetext[3]{\url{https://github.com/harirakul/chemlib}}
\footnotetext[4]{\url{https://github.com/materialsproject/pymatgen}}
\footnotetext[5]{\url{https://github.com/domdfcoding/chemistry_tools}}

\section{Method}
\label{sec:method}

\subsection{Single-Agent Tool-Use Rollout}
\label{sec:rollout}
Figure~\ref{fig:protocol} contrasts the two designs. Where HE-MCTS uses one model to select tools and a separate one to fill their parameters, we use a single policy $\pi_\theta$ that does both. The model talks to a tool server for at most $T=16$ turns. On each turn it writes a short reasoning block inside \texttt{<think>...</think>} and emits one \texttt{<tool\_call>} object of the form \texttt{\{"name": "<lib>/<func>", "arguments": \{...\}\}}. The server executes the call and appends its output as a \texttt{<tool\_result>...</tool\_result>} span, and the model opens its next \texttt{<think>} block using the value actually returned. An episode ends when the model writes a single-line \texttt{Answer:} or exhausts the turn and tool-call budget. Because \texttt{<tool\_result>} spans are produced by the server rather than the policy, we mask them out of the training loss and update only on tokens the model generated itself. We call this procedure, one policy interleaving reasoning with live tool execution in a single generation, our \emph{rollout protocol}. Every model we report, trained or untrained, is run under it, so a difference between rows is a difference in the model and not in the harness. Chains carry this dependency whenever they run, which is why we execute the real tools during training rather than simulating their outputs. Appendix~\ref{app:prompts} gives the prompt, the exact trajectory format and a complete rollout (Figure~\ref{fig:example}) in which a later call's argument is a value an earlier call returned.

\subsection{Supervised Fine-Tuning}
\label{sec:sft}
Reinforcement learning from a cold base model is slow to acquire the tool-calling format, so we optionally warm the policy up with a short round of supervised fine-tuning. We linearise each gold calling chain into the rollout trajectory of \S\ref{sec:rollout}, interleaving a reasoning block, the tool call and its returned result at every step, and set the target final answer to a complete natural-language sentence stating the quantities the question asks for. Training the policy to reproduce these trajectories teaches both the output format and a first approximation of correct tool selection. This checkpoint is the starting point for RL (SFT$\to$RL); we also run RL directly from the base model (RL-only), and find the warm-up is worth at least as much as RL on its own (\S\ref{sec:sftrl}).

\subsection{Programmatic Reward}
\label{sec:reward}
Let $C$ be the ground-truth calling chain, and let a response produce a set of
calls $\hat{C}$ and a final answer $\hat{a}$. Every reward below is computed by a program rather than a learned model, and each is mapped linearly to $[-1,1]$. This is a deliberate departure from CheMatAgent \citep{wu2025CheMatAgent}, whose search is guided by two \emph{learned} critics, a Process Reward Model and an Outcome Reward Model, the latter regressed onto a weighted average of a rule-based score and a GPT-assigned one. Our reward needs no such critics and no LLM judge in the training loop. It depends only on $C$ and is therefore cheap, deterministic, and reproducible. The main variants are the following.
\begin{itemize}
  \item \textbf{$R_\text{ans}$ (value inclusion).} The fraction of the
  ground-truth \emph{return values} that appear verbatim, after normalisation, in the final answer $\hat{a}$. It does not compare $\hat{a}$ with a reference
  answer, but asks only whether the values the tools produced were reported. Full credit requires all of them, and a partial match receives a negative reward.
  \item \textbf{$R_\text{tool}$ and $R_\text{tool}^{F1}$ (tool set).} Recall, or its F1 counterpart, over the set of tool names. F1 penalises extra calls, which discourages the shortcut of invoking everything in the pool.
  \item \textbf{$R_\text{call}$ and $R_\text{call}^{F1}$ (tool and arguments).} As $R_\text{tool}$, but a ground-truth call counts only when its arguments match too, with strings lowercased and floats rounded to six decimals. The $R_\text{call}^{F1}$ variant treats calls as a list rather than a set, so repeats enlarge the denominator and lower precision, discouraging both spamming and padding.
  \item \textbf{$R_\text{hyb}$ and $R_\text{hyb}^{F1}$ (hybrid).} An equal mix of a tool reward and the answer reward,
  $R_\text{hyb} = 0.5\,R_\text{tool} + 0.5\,R_\text{ans}$, with
  $R_\text{hyb}^{F1}$ formed the same way from $R_\text{tool}^{F1}$. The policy is scored both on which calls it made and on whether it reported the values those calls returned.
\end{itemize}

\subsection{Reinforcement Learning}
\label{sec:optim}
We train $\pi_\theta$ with GRPO \citep{shao2024deepseekmath}. For each prompt we sample a group of $n=5$ rollouts, standardize their rewards within the group to obtain advantages (dispensing with PPO's value network), and take a clipped policy-gradient step at learning rate $1\times10^{-6}$. Training runs on the \texttt{slime} post-training stack \citep{slime_github}, where Megatron-LM updates the policy while an SGLang server produces the multi-turn rollouts and the two communicate through slime's data buffer. Both starting points, RL-only and SFT$\to$RL, use identical optimizer settings, so their difference reflects only the warm-up; the backbones and baselines are listed in \S\ref{sec:setup}.


\section{Experimental Setup}
\label{sec:setup}

We adopt CheMatAgent's benchmark, tool exposure and scoring rules throughout, so that a difference in the tables is attributable to training.

\subsection{Benchmark and Backbones}
\label{sec:benchmodels}
\label{sec:retrieval}
We evaluate on the multiple-tool comprehensive-chemistry split of ChemToolBench \citep{wu2025CheMatAgent}, using its \emph{full} test set of $200$ multi-step questions, the same evaluation set CheMatAgent use. A few released gold chains no longer reproduce, because some tools query live services, so we drop those records from the training and development data, leaving $1566$ and $196$ questions (Appendix~\ref{app:refresh}).

We train the same two backbones CheMatAgent report, Qwen-2.5-7B-Instruct \citep{qwen2025qwen25} and Llama-3.1-8B-Instruct \citep{grattafiori2024llama3}, so that within each block of Table~\ref{tab:main} only the training method varies, and add Qwen3-4B \citep{yang2025qwen3} as a scale check, which appears only in Table~\ref{tab:arms}. Each untrained backbone is measured under the same rollout protocol (\S\ref{sec:rollout}), which separates what training contributes from what the protocol itself does.

\subsection{Training and Inference}
The two stages follow \S\ref{sec:sft} (SFT warm-up) and \S\ref{sec:optim} (GRPO), with SFT$\to$RL continuing from the warm-up checkpoint and RL-only starting from the base model. The warm-up runs three epochs at learning rate $10^{-5}$ with batch size $64$ over the $1394$ training questions whose gold chain can be linearised into a supervised trajectory. Reinforcement learning then runs $100$ rollout steps over all $1566$, since a rollout is scored against the gold chain directly and needs no linearised target. At evaluation we decode greedily ($\tau{=}0$), so no difference we report comes from sampling.

\subsection{Evaluation Metrics}
\label{sec:metrics}
We reimplement CheMatAgent's metrics against their released scoring code so that our columns and theirs mean the same thing, micro-averaging over the whole test set as they do. \textbf{Format} is the fraction of emitted tool calls that parse. \textbf{Tool} matches tool names with \emph{list} semantics, so a duplicated call scores as a false positive. \textbf{Param} awards partial credit per parameter value, so a call with two of three arguments right scores $2/3$. \textbf{Return} compares the value a call actually produced against the gold return. \textbf{Pass Rate} is answer-level, computed as CheMatAgent do with their grading prompt verbatim and GPT-4o-mini at temperature $0$. The judge sees the question, the gold answer and the model's answer and returns Yes or No, and we report the fraction of Yes over all $200$ questions, a missing answer counting as No (Appendix~\ref{app:judge}). None of these is involved in training, where the reward is a separate programmatic function (\S\ref{sec:reward}).

\input{table_main}

\section{Results}
\label{sec:results}
\label{sec:mainresults}
Table~\ref{tab:main} compares our models against CheMatAgent's, grouped by backbone so that within a block the base model is held fixed and only the training varies.

\paragraph{Tool selection.}
The clearest result is on tool selection, the task the HE-MCTS policy model exists to perform. On both shared backbones our model takes Tool precision, recall, and F1 outright. On Qwen-2.5-7B we reach $95.76$ Tool F1 against $90.80$ for -M3, the strongest configuration they build on that base; on Llama-3.1-8B, $95.93$ against $92.47$ for -M3 and $92.65$ for their supervised I$^{*}$ model. Because the comparison is within a backbone, \textbf{the same weights, trained differently, select tools better without any search at inference.}

\paragraph{Parameters and returns.}
Returns follow the same pattern as tools. We take all three Return columns on both backbones, $89.39/88.39/88.89$ against -M3's $81.41/80.79/81.10$ on Qwen-2.5-7B and $90.06/89.35/89.70$ against -M2's $88.64/84.07/86.36$ on Llama-3.1-8B. Parameters are more mixed. Param recall and F1 are unavailable for the search rows, so precision is the only Param column on which those rows admit a comparison, and HE-MCTS keeps it on both backbones ($89.32$ and $95.12$ against our $88.35$ and $89.50$). Against the supervised Llama I$^{*}$ model, which does report all three, we lead every Param column. Whatever the search buys here costs four trained models and a tree search at every question.

\paragraph{Answer Pass Rate.}
Pass Rate is the metric that most directly reflects what a user receives, and it
is where the two backbones part company. On Qwen-2.5-7B a single pass reaches
$68.50$, ahead of -M3's $67.32$. On that backbone our model leads every
column HE-MCTS reports comparably except Param precision, so almost everything
the search procedure buys is recovered by training one policy differently. On
Llama-3.1-8B we reach $64.00$ against $72.30$ for -M2, and search retains a clear
advantage. Both scores beat every chain-of-thought baseline in the table,
including GPT-4o-mini ($58.00$) and Claude-3.5-Sonnet ($57.00$), both supervised
models (ChemLLM $53.00$, Llama-3.1-8B-I$^{*}$ $55.00$), and the weaker search
configurations (-M0 $29.19$, Claude-3.5-S-M $57.06$).

One asymmetry behind this comparison runs against us and is worth naming. Our
reward is computed from the gold call chain and never inspects the final answer
(\S\ref{sec:reward}), so nothing in training optimises Pass Rate. What the table
reports is whatever answer quality follows from calling the right tools after a
warm-up on complete answers. CheMatAgent's Outcome Reward Model scores the final
answer directly and is regressed in part onto GPT judgements, and it guides the
search that selects the trajectories their policy is then fine-tuned on. Their
Pass Rate is thus optimised, if indirectly, while ours is a by-product. That our
Qwen model still leads suggests a tool-level objective carries most of the
distance, while our Llama model shows it is not sufficient everywhere.


\section{Analysis}
\label{sec:analysis}

\subsection{What Each Training Stage Contributes}
\label{sec:sftrl}
\input{table_arms}
Table~\ref{tab:arms} runs all four arms (untrained backbone, SFT only, RL only, and the composition) through our rollout protocol on the same $200$ questions, so they differ in training and nothing else. Both RL arms use $100$ rollout steps under $R_\text{call}^{F1}$, and where a supervised stage is present it is the same three-epoch warm-up. Qwen3-4B is added as a scale check.

\paragraph{SFT alone beats RL alone.} SFT-only leads every F1 column and Pass
Rate on all three backbones. On Tool F1 the margin is $1.43$ on Qwen-2.5-7B,
$0.28$ on Llama-3.1-8B and $8.96$ on Qwen3-4B, and on Pass Rate it is $5.50$,
$7.00$ and $8.50$. Every cell RL-only takes is a precision cell, the effect the
next paragraph isolates.

The two margins have different causes. On the tool columns both stages learn from the same gold call chains, but SFT sees them token by token while RL sees only one number per rollout summarising how well it matched, so the same information reaches the model through a much narrower channel. On Pass Rate the problem is not narrowness but absence. Supervised targets end in a complete natural language answer, so SFT is trained on how to \emph{report} a result, while our reward reads the call chain alone and never inspects the answer (\S\ref{sec:reward}). RL has no signal at all for what Pass Rate measures.

\subsection{Reward Design and Reward Hacking}
\label{sec:rewardablation}
\input{table_reward_ablation}
We ablate the reward with the warm-up removed
(Table~\ref{tab:rewardablation}), so a pathology the reward permits appears
undamped by an SFT prior. \textbf{Rewarding recall alone is satisfied by calling everything, and the policy finds that solution.} $R_\text{call}$ never charges for calls outside the gold chain and reaches the highest recall of any arm ($98.25$) by issuing $18.33$ calls per question against a gold average of $3.15$, half of them repeats, which collapses precision to $19.70$ and leaves $97\%$ of episodes exhausting the call budget without answering. One precision term prevents this and nothing further is needed: all three F1 rewards stay within $0.1$ of the gold call count, rarely repeat and always terminate, so the gap between $32.82$ and $95.38$ Tool F1 rests on that term alone. Their remaining differences are second-order, and we use $R_\text{call}^{F1}$ because it takes Param and Return, the columns that check arguments. Scoring the answer as well does not help: $R_\text{hyb}^{F1}$ adds $R_\text{ans}$, which demands each gold return value \emph{verbatim}: only $57\%$ of numeric returns match exactly where $92\%$ are reported to within $1\%$, so it feeds noise rather than signal into training, and $R_\text{hyb}^{F1}$ is the least stable arm.

\subsection{Self-Repair from Execution Feedback}
\label{sec:selfrepair}
Of the three problems in \S\ref{sec:intro}, output-dependent chaining carries a cost the other two problems do not. Because a later call consumes an earlier return, one bad value is inherited by every step after it, even when those steps select and type their own calls correctly. Our protocol executes tools for real in training as well as at test, so a misnamed or wrongly typed call returns an error string rather than a plausible value, and the policy reads it on the next turn. Our models act on this, re-issuing the call with the fault corrected instead of continuing as though it had succeeded. We call this \emph{self-repair from execution feedback}, and supervision cannot be its source, since gold chains contain only calls that succeed.

Figure~\ref{fig:selfrepair} in Appendix~\ref{app:selfrepair} shows a case where one mistyped argument name would otherwise have cost both remaining values, however well those two calls were chosen.

\section{Conclusion}
\label{sec:conclusion}
Chemistry tool learning has been approached with heavy, multi-part systems: a separate planner and executor, trained PRM and ORM critics, and inference-time tree search. We showed that this machinery is not necessary. A single policy that interleaves reasoning and tool calls in one pass, warmed up with a short round of SFT and then optimized with GRPO against a simple programmatic reward, matches or exceeds CheMatAgent's HE-MCTS pipeline on tool, parameter, and return F1 on the same backbones, and is competitive on answer Pass Rate, all at one model invocation per question. Our analysis further shows where each ingredient helps: the supervised warm-up supplies tool recall, RL supplies precision, and list/F1 reward semantics are what keep the policy from hacking the tool score. Verifiable programmatic rewards are enough; no learned reward model or search is required to reach this level.






\section*{Limitations}
\label{sec:limitations}
\paragraph{Scope.} We study one benchmark and one domain, the multiple-call comprehensive-chemistry split of ChemToolBench. To our knowledge no other chemistry benchmark pairs questions with multi-call gold chains, so in-domain generalisation cannot be tested, and whether the recipe transfers to other scientific-tool domains is untested. Its chains run two to six calls and our backbones span $4$--$8$B parameters, so longer horizons and larger models are unevaluated.

\section*{Acknowledgments}
The authors gratefully acknowledge financial support from Air Liquide and the Mitacs Program (IT46383) for this research.
We also acknowledge the computational resources provided by DeltaAI at the National Center for Supercomputing Applications through ACCESS allocation CIS260150 (E.Y., B.L.).


\bibliography{custom}

\appendix


\section{Refreshing stale gold returns}
\label{app:refresh}
\label{sec:refresh}
Some ChemToolBench tools query live services, so for a few records the return stored in the released gold chain is no longer what the tool returns today. A stale chain penalizes a model that calls the right tool with the right arguments and reports what it received, which is the behaviour the metric should reward. The failures are value mismatches in tools whose answer legitimately changes over time, led by \texttt{chemcrow/PatentCheck}, plus a few tools that now raise.
 
A stale chain is a corrupt supervision target, so we drop the affected records from the data we train on, $57$ of $1623$ training and $4$ of $200$ development questions. We drop nothing from the test split.

\section{The chemistry tool pool}
\label{app:toolpool}
The chemistry split is served by $55$ tools across four libraries. Each is invoked by its library-qualified name with typed JSON arguments, and the model sees only the signature, never the implementation. Ten of the $55$ call live web services at request time, which is why a small number of released gold returns no longer reproduce (Appendix~\ref{app:refresh}). Table~\ref{tab:topcalls} lists the calls the gold chains issue most often.

\begin{table}[h]
\centering
\scriptsize
\begin{tabular}{@{}lr@{}}
\toprule
\bf Tool & \bf Calls \\
\midrule
\texttt{chemistrytools/get\_compound\_CID}                  & $732$ \\
\texttt{chemistrytools/get\_compound\_}                     &       \\
\quad\texttt{MolecularWeight\_by\_CID}                      & $616$ \\
\texttt{chem\_lib/get\_empirical\_formula\_}                &       \\
\quad\texttt{by\_percent\_composition}                      & $446$ \\
\texttt{chem\_lib/calculate\_compound\_molar\_mass}         & $438$ \\
\bottomrule
\end{tabular}
\caption{The most frequently called tools in the gold chains of the multiple-tool
chemistry split, counted over train, development and test.}
\label{tab:topcalls}
\end{table}

\section{Prompts and rollout protocol}
\label{app:prompts}
One prompt is used for RL rollouts, for SFT targets and for evaluation, so the
policy never sees a format at test time that it was not trained on. It is
assembled from four parts: the system prompt below, the JSON signatures of the
tools exposed for that query, one worked example, and the user turn.

\paragraph{System prompt.}
\begin{quote}\small\ttfamily
You are a chemistry assistant that solves problems by calling chemistry tools.
Work strictly ONE step at a time:\\
1. Write a brief \texttt{<think>...</think>} block that reasons about ONLY the
next single tool call. Do NOT plan multiple future steps in advance --- you
cannot know what a tool will return until it is actually called, so later steps
must be decided after seeing real \texttt{<tool\_result>} values.\\
2. Emit exactly one \texttt{<tool\_call>...</tool\_call>} JSON block.\\
3. Wait for the \texttt{<tool\_result>}, then start the next
\texttt{<think>} by incorporating what the tool actually returned.\\
When you have enough information, give a COMPLETE final answer on a single line
that states, in a full sentence, every quantity the question asked for, as:
\texttt{Answer: <your complete answer>}.
\end{quote}
The one-step-at-a-time instruction is what forces the interleaving of
\S\ref{sec:rollout}: the model cannot emit a plan for the whole chain, because
each later argument may depend on a value it has not yet seen.

\paragraph{Tools and user turn.} Tool signatures are inserted as JSON objects
inside \texttt{<tools></tools>}, followed by the example, in a chat template
whose role delimiters follow the backbone (Qwen or Llama). The user turn is the
benchmark query with a fixed suffix instructing the model to emit
\texttt{<tool\_call>} blocks and finish with a single \texttt{Answer:} line.

\paragraph{Rollout and supervision.} An episode runs for at most $T=16$ turns.
The server executes each call and appends its output verbatim inside
\texttt{<tool\_result></tool\_result>}; because those spans are produced by the
tool server rather than the policy, they are masked out of the loss and only
model-generated tokens are trained on. Episodes end at the first
\texttt{Answer:} line or when the turn budget is exhausted. SFT targets are gold calling chains linearised into exactly this format (reasoning
block, call, returned value, repeated) and closed with a complete natural-language
answer, so the supervised and RL formats are identical.
\input{fig_example}

\section{Answer-level judging}
\label{app:judge}
Pass Rate is computed exactly as \citet{wu2025CheMatAgent} compute it, reusing
their grading prompt verbatim:
\begin{quote}\small\ttfamily
Please evaluate whether the given answer to the question is correct according to
the gold answer. Return "Yes" if it is correct, otherwise return "No".\\[2pt]
Question:\\ \{query\}\\[2pt]
Gold Answer:\\ \{gold\}\\[2pt]
Given Answer:\\ \{pred\}\\[2pt]
Whether the given answer is correct:
\end{quote}
The judge is GPT-4o-mini at temperature $0$. \texttt{\{gold\}} is their released
natural-language gold answer for the question, not the raw tool returns, matching
the reference their own \texttt{response\_generation.py} supplies;
\texttt{\{pred\}} is the text after the final \texttt{Answer:} marker in the
rollout. Pass Rate is the fraction of \texttt{Yes} verdicts over all $200$ test
questions, with a missing or empty answer counted as \texttt{No}.

\section{A self-repair rollout}
\label{app:selfrepair}
Figure~\ref{fig:selfrepair} is the unedited training rollout discussed in \S\ref{sec:selfrepair}, in which the model reads an execution error and re-issues the failed call corrected. Gold chains contain only calls that succeed, so this behaviour cannot have been imitated from supervision.
\input{fig_selfrepair}

\end{document}

%% file: fig_protocol.tex
\begin{figure*}[t]
\centering
\definecolor{cbBlue}{RGB}{86,180,233}
\definecolor{cbViolet}{RGB}{117,112,179}
\definecolor{cbGreen}{RGB}{0,158,115}
\definecolor{cbGrey}{RGB}{130,130,130}
\definecolor{cbPink}{RGB}{204,121,167}
\definecolor{cbOrange}{RGB}{230,159,0}
\definecolor{cbVerm}{RGB}{213,94,0}
\resizebox{\textwidth}{!}{%
\begin{tikzpicture}[
  >=Stealth, font=\small,
  nd/.style={circle, draw=black!55, minimum size=4.4mm, inner sep=0pt, fill=white},
  ndsel/.style={nd, fill=cbOrange!45, draw=cbOrange!80!black},
  ndpr/.style={nd, densely dashed, draw=black!25},
  chip/.style={rounded corners=2pt, draw=black!55, fill=black!4,
               inner xsep=4pt, inner ysep=2.5pt, align=center, font=\scriptsize},
  chipc/.style={chip, draw=cbVerm!70!black, fill=cbVerm!8},
  stbox/.style={rounded corners=3pt, draw=black!55, fill=black!4, align=center,
                font=\scriptsize, inner xsep=5pt, inner ysep=4pt},
  llm/.style={rounded corners=3pt, draw=cbBlue!80!black, line width=0.8pt,
              fill=cbBlue!18, align=center, font=\small\bfseries, inner ysep=5pt},
  srvbox/.style={rounded corners=3pt, draw=black!55, line width=0.8pt,
                 fill=black!8, align=center, font=\small\bfseries, inner ysep=5pt},
  qbox/.style={rounded corners=2pt, draw=black!55, fill=white, align=center,
               font=\small, inner xsep=5pt, inner ysep=4pt},
  qbs/.style={qbox, font=\scriptsize, inner xsep=4pt, inner ysep=3pt},
  bar/.style={rounded corners=1pt, minimum height=3.4mm, inner sep=0pt},
  th/.style={bar, fill=cbViolet!40, draw=cbViolet!85!black, minimum width=19mm},
  tc/.style={bar, fill=cbGreen!30, draw=cbGreen!80!black, minimum width=9mm},
  tr/.style={bar, fill=cbGrey!22,  draw=cbGrey!95!black, densely dashed,
             minimum width=9mm},
  fa/.style={bar, fill=cbPink!38,  draw=cbPink!85!black,  minimum width=19mm},
  old/.style={opacity=0.5},
  stepbg/.style={rounded corners=3pt, draw=black!25, fill=black!4, inner sep=4pt},
  rw/.style={rounded corners=2pt, draw=black!60, fill=white, align=center,
             font=\small, inner xsep=6pt, inner ysep=4pt},
  ttl/.style={font=\bfseries},
  stl/.style={font=\scriptsize\bfseries, text=black!70},
  ann/.style={font=\scriptsize\itshape, text=black!60, align=center},
  ar/.style={->, draw=black!60, line width=0.6pt},
  arg/.style={->, draw=cbGreen!80!black, line width=0.8pt},
  ars/.style={->, draw=black!50, line width=0.8pt},
  ard/.style={->, draw=cbOrange!85!black, line width=0.9pt, densely dotted},
]

\node[ttl, anchor=west] at (-0.25,9.42) {(a) CheMatAgent: HE-MCTS --- four models, search at training and test time};

\node[qbs]   (qa) at (0.45,8.95) {Question};
\node[nd]    (r)  at (2.15,8.95) {};
\node[ndsel] (c1) at (1.35,8.30) {};
\node[nd]    (c2) at (2.15,8.30) {};
\node[ndpr]  (c3) at (2.95,8.30) {};
\node[ndsel] (g1) at (1.00,7.65) {};
\node[ndpr]  (g2) at (1.70,7.65) {};
\node[nd]    (g3) at (2.55,7.65) {};
\node[ndpr]  (g4) at (3.25,7.65) {};
\node[qbs]   (aa) at (1.00,6.95) {Answer};
\draw[ar]  (qa) -- (r);
\draw[ars] (r) -- (c1); \draw[ars] (r) -- (c2); \draw[ars] (r) -- (c3);
\draw[ars] (c1) -- (g1); \draw[ars] (c1) -- (g2);
\draw[ars] (c2) -- (g3); \draw[ars] (c2) -- (g4);
\draw[ar]  (g1) -- (aa);
\node[ann, anchor=west] at (0.30,6.55)
  {one search iteration: select $\to$ expand $\to$ evaluate $\to$ simulate $\to$ back-propagate};

\node[chip]  (pm)  at (6.60,8.95) {\textbf{Policy model} $p$ --- proposes $k$ calls per node};
\node[chip]  (em)  at (6.60,8.30) {\textbf{Execution model} $u$ --- fills parameters, retries};
\node[chipc] (prm) at (6.60,7.65) {\textbf{PRM} --- learned critic, scores each step};
\node[chipc] (orm) at (6.60,7.00) {\textbf{ORM} --- learned critic, scores the answer};
\draw[decorate,decoration={brace,amplitude=4pt}] (3.78,6.84) -- (3.78,9.12);

\node[stbox] (st) at (12.30,8.45) {\textbf{Self-training loop}\\ HE-MCTS trajectories retrain $p$\\ and train both critics};
\draw[ar, densely dashed] (9.45,8.45) -- (st.west);
\node[ann] at (12.30,7.15) {the number of model calls grows\\ with the size of the tree};

\draw[black!35, densely dashed] (-0.25,6.25) -- (14.90,6.25);

\node[ttl, anchor=west] at (-0.25,5.78) {(b) Ours: single-agent rollout with interleaved tool execution};
\node[qbox] (q) at (0.6,3.30) {Question};
\node[llm, minimum width=98mm] (pol) at (6.4,5.05) {Policy LLM $\pi_\theta$};
\draw[ar] (q.north) |- (pol.west);

\node[th] (a1) at (2.5,3.90) {};
\node[tc] (a2) at (2.0,3.45) {};
\node[tr] (a3) at (3.0,3.45) {};

\node[th, old] (b1) at (5.8,3.90) {};
\node[tc, old] (b2) at (5.3,3.45) {};
\node[tr, old] (b3) at (6.3,3.45) {};
\node[th]      (b4) at (5.8,3.00) {};
\node[tc]      (b5) at (5.3,2.55) {};
\node[tr]      (b6) at (6.3,2.55) {};

\node[font=\large] (dots) at (8.0,3.90) {$\cdots$};

\node[th, old] (d1) at (10.2,3.90) {};
\node[tc, old] (d2) at ( 9.7,3.45) {};
\node[tr, old] (d3) at (10.7,3.45) {};
\node[th, old] (d4) at (10.2,3.00) {};
\node[tc, old] (d5) at ( 9.7,2.55) {};
\node[tr, old] (d6) at (10.7,2.55) {};
\node[th]      (d7) at (10.2,2.10) {};
\node[fa]      (d8) at (10.2,1.65) {};

\begin{scope}[on background layer]
  \node[stepbg, fit=(a1)(a2)(a3)] (S1) {};
  \node[stepbg, fit=(b1)(b5)(b6)] (S2) {};
  \node[stepbg, fit=(d1)(d7)(d8)] (S3) {};
\end{scope}
\node[stl, anchor=south west] at ([xshift=1pt,yshift=1.5pt]S1.north west) {Step 1};
\node[stl, anchor=south west] at ([xshift=1pt,yshift=1.5pt]S2.north west) {Step 2};
\node[stl, anchor=south west] at ([xshift=1pt,yshift=1.5pt]S3.north west) {Step $T$};

\draw[ar] (S1.east |- a1) -- (S2.west |- b1);
\draw[ar] (S2.east |- b1) -- (dots.west);
\draw[ar] (dots.east) -- (S3.west |- d1);
\draw[ar] (pol.south -| S1.north) -- (S1.north);
\draw[ar] (pol.south -| S2.north) -- (S2.north);
\draw[ar] (pol.south -| S3.north) -- (S3.north);

\node[srvbox, minimum width=80mm] (srv) at (6.2,0.70) {Chemistry tool server \;\scriptsize(real execution)};
\draw[arg] (a2.south) -- (a2.south |- srv.north);
\draw[ars] (a3.south |- srv.north) -- (a3.south);
\draw[arg] (b5.south) -- (b5.south |- srv.north);
\draw[ars] (b6.south |- srv.north) -- (b6.south);

\draw[ard, shorten >=1pt] (a3.east) to[out=-30,in=160] (b5.west);
\draw[ard, shorten >=1pt] (b6.east) to[out=-35,in=215] (d5.west);
\node[ann, text=cbOrange!75!black] at (4.05,1.85)
  {output dependence:\\ a return becomes the\\ next call's argument};

\node[rw] (rew) at (13.30,2.55) {Reward};
\node[rw] (adv) at (13.30,5.05) {Advantage};
\draw[ar] (S3.east) -- (rew.west);
\draw[ar] (rew) -- (adv);
\draw[ar] (adv.west) -- (pol.east);
\node[ann] at (13.30,1.78) {computed by program\\ from the gold call chain;\\ no learned critic};

\node[th, minimum width=8mm] at (11.90,0.95) {};
\node[anchor=west, font=\footnotesize] at (12.35,0.95) {\texttt{<think>}};
\node[tc, minimum width=8mm] at (11.90,0.60) {};
\node[anchor=west, font=\footnotesize] at (12.35,0.60) {\texttt{<tool\_call>}};
\node[tr, minimum width=8mm] at (11.90,0.25) {};
\node[anchor=west, font=\footnotesize] at (12.35,0.25) {\texttt{<tool\_result>}};
\node[fa, minimum width=8mm] at (11.90,-0.10) {};
\node[anchor=west, font=\footnotesize] at (12.35,-0.10) {\texttt{<Answer>}};

\end{tikzpicture}}
\caption{\textbf{(a)} CheMatAgent's HE-MCTS \citep{wu2025CheMatAgent} splits tool
selection and parameter filling across two models and searches a tree under two
\emph{learned} critics --- a Process Reward Model (PRM) over steps and an Outcome
Reward Model (ORM) over answers, the latter trained on a GPT-4o assessment blended
with a rule-based check. Four models are trained and deployed, and model calls grow
with the tree. \textbf{(b)} One policy $\pi_\theta$ interleaves reasoning with live
tool execution in a single generation, so a question costs one model and one LLM
call. Context accumulates until step $T$ ends in an answer; faded rows are earlier
context, and \texttt{<tool\_result>} spans (dashed) come from the server rather than
the policy. A program then scores the trajectory against the gold call chain --- no
learned critic, no search (\S\ref{sec:rollout}).}
\label{fig:protocol}
\end{figure*}

%% file: table_main.tex
\begin{table*}[t]
\centering
\footnotesize
\setlength{\tabcolsep}{3.4pt}
\begin{tabular}{l c ccc ccc ccc c}
\toprule
 & & \multicolumn{3}{c}{Tool} & \multicolumn{3}{c}{Param} & \multicolumn{3}{c}{Return} & Pass \\
\cmidrule(lr){3-5}\cmidrule(lr){6-8}\cmidrule(lr){9-11}
Model & Format & P & R & F1 & P & R & F1 & P & R & F1 & Rate \\
\midrule
GPT-4o-mini (CoT)        & 99.83 & 83.22 & 78.86 & 80.98 & 76.04 & 72.19 & 74.06 & 77.05 & 73.13 & 75.04 & 58.00 \\
Claude-3.5-S (CoT)       & 97.42 & 83.64 & 85.37 & 84.50 & 76.03 & 77.53 & 76.77 & 74.96 & 78.54 & 76.71 & 57.00 \\
ChemLLM$^{*}$ (SFT)      & 98.10 & 76.84 & 88.08 & 82.07 & 68.22 & 80.20 & 73.72 & 68.84 & 80.45 & 74.19 & 53.00 \\
GPT-4o-mini-M (HE-MCTS)  & /     & 85.06 & 83.57 & 84.31 & 88.47 & / & / & 75.97 & 74.64 & 75.30 & 62.30 \\
Claude-3.5-S-M (HE-MCTS) & /     & 89.80 & 86.27 & 88.00 & 86.36 & / & / & 77.55 & 74.51 & 76.00 & 57.06 \\
\midrule
\multicolumn{12}{l}{\emph{Base: Qwen-2.5-7B-Instruct}}\\
\quad CheMatAgent-I (no FT)      & 51.25 & 64.15 & 41.81 & 50.63 & 56.48 & 37.36 & 44.97 & 29.25 & 37.20 & 32.75 & 22.50 \\
\quad CheMatAgent-M3 (HE-MCTS)   & /     & 91.14 & 90.45 & 90.80 & \bf 89.32 & / & / & 81.41 & 80.79 & 81.10 & 67.32 \\
\quad \bf Ours (SFT$\to$RL)    & \bf 100.00 & \bf 96.30 & \bf 95.23 & \bf 95.76 & 88.35 & \bf 87.36 & \bf 87.85 & \bf 89.39 & \bf 88.39 & \bf 88.89 & \bf 68.50 \\
\midrule
\multicolumn{12}{l}{\emph{Base: Llama-3.1-8B-Instruct}}\\
\quad CheMatAgent-I (no FT)      & 75.99 & 59.95 & 38.31 & 46.75 & 52.98 & 33.71 & 41.20 & 40.83 & 34.34 & 37.31 & 15.00 \\
\quad CheMatAgent-I$^{*}$ (SFT)  & 99.20 & 93.10 & 92.21 & 92.65 & 83.85 & 83.15 & 83.50 & 85.03 & 84.90 & 84.96 & 55.00 \\
\quad CheMatAgent-M0 (HE-MCTS)   & /     & 75.45 & 78.45 & 76.92 & 85.25 & / & / & 65.09 & 67.68 & 66.36 & 29.19 \\
\quad CheMatAgent-M1 (HE-MCTS)   & /     & 87.74 & 87.32 & 87.53 & 85.92 & / & / & 75.81 & 75.44 & 75.62 & 69.50 \\
\quad CheMatAgent-M2 (HE-MCTS)   & /     & 93.18 & 88.39 & 90.79 & \bf 95.12 & / & / & 88.64 & 84.07 & 86.36 & \bf 72.30 \\
\quad CheMatAgent-M3 (HE-MCTS)   & /     & 93.22 & 91.73 & 92.47 & 92.36 & / & / & 86.09 & 84.72 & 85.41 & 72.20 \\
\quad \bf Ours (SFT$\to$RL)    & \bf 100.00 & \bf 96.31 & \bf 95.55 & \bf 95.93 & 89.50 & \bf 88.62 & \bf 89.06 & \bf 90.06 & \bf 89.35 & \bf 89.70 & 64.00 \\
\bottomrule
\end{tabular}
\caption{Main results on the multiple-tool comprehensive-chemistry benchmark,
grouped by backbone so that within a block only the training method varies. All
rows use the same $200$ test questions under CheMatAgent's evaluation regime, and
\textbf{bold marks the best value within each shared-backbone block}. On both
backbones our single policy takes every Tool and Return column with no tree
search, and on Qwen-2.5-7B it also takes Pass Rate. $^{*}$: fine-tuned on the
comprehensive-chemistry split. ``/'': value unavailable.}
\label{tab:main}
\end{table*}

%% file: table_arms.tex
\begin{table*}[t]
\centering
\footnotesize
\setlength{\tabcolsep}{4pt}
\begin{tabular}{l ccc ccc ccc c}
\toprule
 & \multicolumn{3}{c}{Tool} & \multicolumn{3}{c}{Param} & \multicolumn{3}{c}{Return} & Pass \\
\cmidrule(lr){2-4}\cmidrule(lr){5-7}\cmidrule(lr){8-10}
Training & P & R & F1 & P & R & F1 & P & R & F1 & Rate \\
\midrule
\multicolumn{11}{l}{\emph{Qwen-2.5-7B-Instruct}}\\
\quad none (base)        & 85.09 & 86.17 & 85.62 & 75.98 & 76.40 & 76.19 & 76.45 & 77.42 & 76.94 & 54.00 \\
\quad SFT only           & 93.91 & 93.16 & 93.54 & 85.98 & 85.25 & 85.61 & 87.50 & 86.80 & 87.15 & 65.50 \\
\quad RL only            & 94.33 & 89.98 & 92.11 & 85.86 & 82.72 & 84.26 & 88.67 & 84.58 & 86.57 & 60.00 \\
\quad \bf SFT$\to$RL     & \bf 96.30 & \bf 95.23 & \bf 95.76 & \bf 88.35 & \bf 87.36 & \bf 87.85 & \bf 89.39 & \bf 88.39 & \bf 88.89 & \bf 68.50 \\
\midrule
\multicolumn{11}{l}{\emph{Llama-3.1-8B-Instruct}}\\
\quad none (base)        & 36.70 & 84.90 & 51.25 & 20.82 & 54.63 & 30.16 & 25.50 & 58.98 & 35.60 & 24.50 \\
\quad SFT only           & 94.90 & 94.59 & 94.75 & 87.76 & 87.64 & 87.70 & 88.36 & 88.08 & 88.22 & \bf 69.50 \\
\quad RL only            & 95.31 & 93.64 & 94.47 & 86.86 & 85.39 & 86.12 & 87.86 & 86.33 & 87.09 & 62.50 \\
\quad \bf SFT$\to$RL     & \bf 96.31 & \bf 95.55 & \bf 95.93 & \bf 89.50 & \bf 88.62 & \bf 89.06 & \bf 90.06 & \bf 89.35 & \bf 89.70 & 64.00 \\
\midrule
\multicolumn{11}{l}{\emph{Qwen3-4B}}\\
\quad none (base)        & 87.88 & 73.77 & 80.21 & 80.84 & 67.56 & 73.60 & 83.14 & 69.79 & 75.89 & 53.00 \\
\quad SFT only           & 93.33 & \bf 93.48 & 93.41 & 84.59 & 84.83 & 84.71 & 86.83 & 86.96 & 86.89 & 65.50 \\
\quad RL only            & 93.10 & 77.27 & 84.45 & 87.90 & 72.47 & 79.45 & 89.27 & 74.09 & 80.97 & 57.00 \\
\quad \bf SFT$\to$RL     & \bf 96.30 & 95.07 & \bf 95.68 & \bf 88.35 & \bf 87.36 & \bf 87.85 & \bf 90.02 & \bf 88.87 & \bf 89.44 & \bf 68.00 \\
\bottomrule
\end{tabular}
\caption{Contribution of each training stage. All arms use our rollout protocol,
the same $200$ test questions and the reward $R_\text{call}^{F1}$, so they differ only
in training (\S\ref{sec:sftrl}).}
\label{tab:arms}
\end{table*}

%% file: table_reward_ablation.tex
\begin{table*}[t]
\centering
\footnotesize
\setlength{\tabcolsep}{3.4pt}
\begin{tabular}{l ccc ccc ccc ccc}
\toprule
 & \multicolumn{3}{c}{Tool} & \multicolumn{3}{c}{Param} & \multicolumn{3}{c}{Return} & \multicolumn{3}{c}{behaviour} \\
\cmidrule(lr){2-4}\cmidrule(lr){5-7}\cmidrule(lr){8-10}\cmidrule(lr){11-13}
Reward & P & R & F1 & P & R & F1 & P & R & F1 & calls & dup. & n-term \\
\midrule
\multicolumn{13}{l}{\emph{extra calls unpenalised}}\\
$R_\text{call}$ (recall, name$+$args)
 & 19.70 & \bf 98.25 & 32.82 & 16.13 & \bf 85.96 & 27.17 & 17.88 & \bf 89.19 & 29.79
 & 18.33 & 49.6\% & 97.0\% \\
\midrule
\multicolumn{13}{l}{\emph{precision term added}}\\
$R_\text{tool}^{F1}$ (set F1, names)
 & \bf 95.53 & 95.23 & \bf 95.38 & 82.87 & 83.57 & 83.22 & 85.33 & 85.06 & 85.19
 & 3.13 & \bf 0.6\% & \bf 0.0\% \\
$R_\text{call}^{F1}$ (list F1, $+$args)
 & 94.33 & 89.98 & 92.11 & \bf 85.86 & 82.72 & \bf 84.26 & \bf 88.67 & 84.58 & \bf 86.57
 & \bf 3.06 & 1.1\% & \bf 0.0\% \\
$R_\text{hyb}^{F1}$ (tool F1 $+$ value incl.)
 & 93.81 & 93.96 & 93.88 & 82.13 & 83.29 & 82.71 & 85.56 & 85.69 & 85.62
 & 3.16 & 3.9\% & \bf 0.0\% \\
\bottomrule
\end{tabular}
\caption{Reward ablation on Qwen-2.5-7B, trained for $100$ steps \emph{without} a
warm-up so the reward is the only signal. ``calls'' is the mean tool calls per
question against a gold average of $3.15$, ``dup.'' the share repeating a tool
already used, and ``n-term'' the share of episodes exhausting the $16$-call budget
without answering (\S\ref{sec:rewardablation}).}
\label{tab:rewardablation}
\end{table*}

%% file: fig_example.tex
\newcommand{\srvline}[1]{%
  \colorbox{black!8}{\makebox[\dimexpr\linewidth-2\fboxsep\relax][l]{#1}}}
\begin{figure}[t]
\centering
\setlength{\fboxsep}{1.5pt}
\begin{tikzpicture}[font=\scriptsize\ttfamily]
\node[rounded corners=2pt, draw=black!55, fill=white, inner sep=5pt,
      text width=\dimexpr\columnwidth-16pt\relax, align=left] (box) {%
\normalfont\footnotesize\textbf{Query}\ \ \textit{(multiple\_1991)}\\[2pt]
{\normalfont\small Please retrieve the PubChem Compound Identifier (CID) for
the chemical compound '(5-Bromo-2-methoxyphenyl)methanol', and subsequently,
obtain the molecular weight of this compound using the retrieved CID.}\\[4pt]
{\color{black!40}\rule{\linewidth}{0.4pt}}\\[3pt]
<think>\\
{\color{black!70}We need to retrieve the PubChem Compound}\\
{\color{black!70}Identifier for '(5-Bromo-2-methoxyphenyl)-}\\
{\color{black!70}methanol', so we call get\_compound\_CID}\\
{\color{black!70}with the compound name.}\\
</think>\\[2pt]
<tool\_call>\\
\{"name": "chemistrytools/get\_compound\_CID",\\
\ "arguments": \{"compound": "(5-Bromo-2-\\
\ methoxyphenyl)methanol"\}\}\\
</tool\_call>\\[3pt]
\srvline{<tool\_result>\ \ {\normalfont\itshape\color{black!55}from server}}\\
\srvline{\textbf{612298}}\\
\srvline{</tool\_result>}\\[3pt]
<think>\\
{\color{black!70}With the PubChem CID (\textbf{612298}) in hand, we}\\
{\color{black!70}call get\_compound\_MolecularWeight\_by\_CID}\\
{\color{black!70}to obtain its molecular weight.}\\
</think>\\[2pt]
<tool\_call>\\
\{"name": "chemistrytools/get\_compound\_\\
\ MolecularWeight\_by\_CID",\\
\ "arguments": \{"compound\_CID": \textbf{612298}\}\}\\
</tool\_call>\\[3pt]
\srvline{<tool\_result>\ \ {\normalfont\itshape\color{black!55}from server}}\\
\srvline{\textbf{217.06}}\\
\srvline{</tool\_result>}\\[3pt]
{\color{teal!65!black}Answer: The molecular weight of '(5-Bromo-2-}\\
{\color{teal!65!black}methoxyphenyl)methanol' is \textbf{217.06} g/mol.}
};
\end{tikzpicture}
\caption{A complete, unedited rollout from our Qwen-2.5-7B SFT$\to$RL model.
Shaded spans are returned by the tool server and masked from the loss; everything
else is generated by the policy. The CID \texttt{612298} appears nowhere in the
query and cannot be guessed, so the second call can only be written by reading it
back out of the first call's result. This rollout matches the gold chain exactly.}
\label{fig:example}
\end{figure}

%% file: fig_selfrepair.tex
\newcommand{\repairline}[1]{%
  \colorbox{black!8}{\makebox[\dimexpr\linewidth-2\fboxsep\relax][l]{#1}}}
\begin{figure}[!htbp]
\centering
\setlength{\fboxsep}{1.5pt}
\begin{tikzpicture}[font=\scriptsize\ttfamily]
\node[rounded corners=2pt, draw=black!55, fill=white, inner sep=5pt,
      text width=\dimexpr\columnwidth-16pt\relax, align=left] (box) {%
\normalfont\footnotesize\textbf{Query}\\[2pt]
{\normalfont\small Please retrieve the Canonical SMILES representation for the
chemical compound with PubChem Compound Identifier (CID) 258. Once you have the
SMILES, calculate the log of the partition coefficient (LogP) and the
Quantitative Estimate of Druglikeness (QED) for the compound.}\\[4pt]
{\color{black!40}\rule{\linewidth}{0.4pt}}\\[3pt]
<tool\_call>\\
\{"name": "chemistrytools/convert\_compound\_\\
\ CID\_to\_SMILES",\\
\ "arguments": \{{\color{red!55!black}\textbf{"compound CID"}}: 258\}\}\\
</tool\_call>\\[3pt]
\repairline{<tool\_result>\ \ {\normalfont\itshape\color{black!55}from server}}\\
\repairline{\color{red!55!black}\textbf{Error: argument binding failed for}}\\
\repairline{\color{red!55!black}chemistrytools/convert\_compound\_CID\_to\_}\\
\repairline{\color{red!55!black}SMILES: got an unexpected keyword}\\
\repairline{\color{red!55!black}argument 'compound CID'}\\
\repairline{</tool\_result>}\\[3pt]
{\color{blue!45!black}\itshape\% self-repair: correct the argument name}\\[1pt]
<tool\_call>\\
\{"name": "chemistrytools/convert\_compound\_\\
\ CID\_to\_SMILES",\\
\ "arguments": \{{\color{blue!45!black}\textbf{"compound\_CID"}}: 258\}\}\\
</tool\_call>\\[3pt]
\repairline{<tool\_result>}\\
\repairline{\textbf{CN(CCC(CC(=O)NC1C=CC(OC1C(=O)O)N2C=CC}}\\
\repairline{\textbf{(=NC2=O)N)N)C(=N)N}}\\
\repairline{</tool\_result>}\\[3pt]
{\color{blue!45!black}\itshape\% this value is the argument of both calls below}\\[1pt]
<tool\_call>\\
\{"name": "cactus/CalculateLogP",\\
\ "arguments": \{"compound\_SMILES": \textbf{<SMILES>}\}\}\\
</tool\_call>\\[3pt]
\repairline{<tool\_result>}\\
\repairline{\textbf{-2.2187299999999963}}\\
\repairline{</tool\_result>}\\[3pt]
<tool\_call>\\
\{"name": "cactus/CalculateQED",\\
\ "arguments": \{"compound\_SMILES": \textbf{<SMILES>}\}\}\\
</tool\_call>\\[3pt]
\repairline{<tool\_result>}\\
\repairline{\textbf{0.150577431129479}}\\
\repairline{</tool\_result>}\\[3pt]
{\color{teal!65!black}Answer: The Canonical SMILES representation}\\
{\color{teal!65!black}for the chemical compound with PubChem}\\
{\color{teal!65!black}Compound Identifier (CID) 258 is \textbf{CN(CCC(CC}}\\
{\color{teal!65!black}\textbf{(=O)NC1C=CC(OC1C(=O)O)N2C=CC(=NC2=O)N)N)}}\\
{\color{teal!65!black}\textbf{C(=N)N}, its log of the partition}\\
{\color{teal!65!black}coefficient (LogP) is \textbf{-2.2187}, and the}\\
{\color{teal!65!black}Quantitative Estimate of Druglikeness (QED)}\\
{\color{teal!65!black}is \textbf{0.1506}.}
};
\end{tikzpicture}
\caption{Self-repair from execution feedback, in an unedited Qwen-2.5-7B RL-only training rollout; shaded spans are returned by the server. The model writes an argument name with a space, reads the binding error, and re-issues the call corrected. That recovered SMILES is the argument of both later calls, so one uncorrected error would have cost two more values. \texttt{<SMILES>} abbreviates the string, which the rollout repeats in full.}
\label{fig:selfrepair}
\end{figure}